\documentclass{article}

\usepackage[utf8]{inputenc}
\usepackage[T1]{fontenc}
\usepackage{authblk}
\usepackage{amssymb}
\usepackage{amsmath}
\usepackage{amsthm}
\usepackage{adjustbox}
\usepackage{bbold}
\usepackage{physics}
\usepackage{paralist}
\usepackage[misc]{ifsym}
\usepackage{epsfig}
\usepackage{algorithm,algpseudocode}

\usepackage{tikz,pgf}
\usepackage{geometry}
\usetikzlibrary{arrows}
\usepackage{pgfplots}
\pgfplotsset{compat=1.5}
\definecolor{BLUE}{rgb}{0.30196078431372547,0.30196078431372547,1}
\definecolor{RED}{rgb}{1,0,0}

\DeclareMathOperator*{\argmax}{arg\,max}
\DeclareMathOperator*{\argmin}{arg\,min}

\newcommand{\intint}[2]{\left[\!\left[#1,#2\right]\!\right]}
\newcommand{\card}{\text{card}}

\newcommand{\1}{\mathbb{1}}
\newcommand{\E}{\mathbb{E}}
\newcommand{\M}{\mathcal{M}}
\newcommand{\N}{\mathbb{N}}
\renewcommand{\P}{\mathbb{P}}
\newcommand{\R}{\mathbb{R}}

\newcommand{\Z}{\mathcal{Z}}

\newcommand{\bdelta}{\overline{\delta}}
\newcommand{\bL}{\overline{L}}
\newcommand{\bmu}{\overline{\mu}}
\newcommand{\bnu}{\overline{\nu}}
\newcommand{\bpi}{\overline{\pi}}
\newcommand{\bPi}{\overline{\Pi}}
\newcommand{\bP}{\overline{P}}

\newcommand{\halpha}{\widehat{\alpha}}
\newcommand{\hdelta}{\widehat{\delta}}
\newcommand{\hL}{\widehat{L}}
\newcommand{\hmu}{\widehat{\mu}}
\newcommand{\hnu}{\widehat{\nu}}
\newcommand{\hpi}{\widehat{\pi}}
\newcommand{\hPi}{\widehat{\Pi}}
\newcommand{\hP}{\widehat{P}}
\newcommand{\hrho}{\widehat{\rho}}
\newcommand{\hz}{\widehat{z}}

\newcommand{\Note}[1]{\textcolor{blue}{\textit{(#1)}}}

\theoremstyle{plain}
\newtheorem{thm}{Theorem}[section]

\newtheorem{rem}[thm]{Remark}
\newtheorem{ass}{Assumption}

\theoremstyle{definition}

\theoremstyle{remark}

\usepackage[justification=centering]{caption}
\DeclareCaptionFormat{sanslabel}{#3}%

\title{Model-free algorithms for fast node clustering in SBM type graphs and application to social role inference in animals}
\author{Bertrand Cloez$^1$, Adrien Cotil$^{2}$, Jean-Baptiste Menassol$^3$, Nicolas Verzelen$^1$}
\affil{\footnotesize $^1$ UMR MISTEA, Univ Montpellier, INRAE, Institut Agro, 34060 Montpellier, France.}
\affil{\footnotesize $^2$ Laboratoire Jacques-Louis Lions, Sorbonne Université, 75005 Paris, France.}
\affil{\footnotesize $^3$ UMR SELMET, Univ Montpellier, Institut Agro, CIRAD, INRAE,  34060 Montpellier, France.}

\date{}

\usepackage{ulem}
\def\cmb#1{\marginpar{\raggedright\tiny{\textcolor{purple}{Bertrand: #1}}}}
\def\bc#1{\textcolor{purple}{#1}}

\begin{document}

\maketitle

\begin{abstract}
We propose a novel family of model-free algorithms for node clustering and parameter inference in graphs generated from the Stochastic Block Model (SBM), a fundamental framework in community detection. Drawing inspiration from the Lloyd algorithm for the $k$-means problem, our approach extends to SBMs with general edge weight distributions. We establish the consistency of our estimator under a natural identifiability condition. Through extensive numerical experiments, we benchmark our methods against state-of-the-art techniques, demonstrating significantly faster computation times with the lower order of estimation error. Finally, we validate the practical relevance of our algorithms by applying them to empirical network data from behavioral ecology.

\tableofcontents
\end{abstract}

\newpage

\section{Introduction}

Graphs have become extremely useful for representing a wide variety of systems in different contexts: biological, social, information... A basic attempt to study them may consist in partitioning the vertices of a graph into clusters that are more densely connected; that is commonly called \textit{community detection} or \textit{graph clustering}; see for instance \cite{fortunato2010community, abbe2018community}. Community detection and clustering are central problems in machine learning and data science. 
In particular, the \textit{stochastic block model} (SBM) \cite{nowicki2001estimation,lee2019review} has been widely used as a canonical model for community detection and as a building block for clustering with more structural assumptions. In its most general form, the SBM corresponds to a randomly weighted graph model where each node has an unobserved label and the probability of observing a given edge between two nodes depends only on the labels of the nodes under consideration. In its simplest form, we recover the Erdös-Renyi random graph model \cite{erdos1959random}, but many more complex variants have been proposed. Among many variants, they may include covariates \cite{mariadassou2010uncovering}, continuous-valued labels \cite{matias2014modeling}, mixed-membership~\cite{airoldi2008mixed}, multi-layer networks~\cite{han2015consistent, valles2016multilayer} or time-varying parameters \cite{matias2017statistical}.

In all these settings, neither the labels are observed nor the conditional probability of observing an edge given the labels is known. One main statistical question posed by this model is then to recover the labels of the nodes and the parameters of the model from an observation of this graph. 
The parameter inference for this model has been studied in \cite{celisse2012consistency} where it is proved that the Maximum Likelihood Estimator (MLE) is consistent and in \cite{bickel2013asymptotic} where it is proved that it is asymptotically normal. However, computing the MLE includes a difficult optimization problem. To overcome this problem, variational EM (VEM) methods \cite{blei2017variational} are often used \cite{chabert2024learning}. The associated estimator remains consistent \cite{celisse2012consistency} and asymptotically normal \cite{zhang2020theoretical}. This powerful method is implemented in \cite{Chiquet2024Rsbm}. In the Bayesian context, several MCMC algorithms have been introduced \cite{mcdaid2013improved,peixoto2014efficient, choi2012stochastic}, some of which are based on the Variational Bayes \cite{wang2018frequentist}, $i.e.$ the Bayesian equivalent of VEM. Variational EM algorithms have also been developed for parameter inference in more complex models, as in \cite{tabouy2020variational}, where a variant of SBM with unobserved data is studied, or in \cite{chabert2021stochastic}, where the observed graph is assumed to be organized in several layers.

If one focuses on the label recovery problem (or equivalently the node clustering), there has been a long line of research (see in particular~\cite{abbe2018community}) towards developing optimal and polynomial-time algorithms, which includes semi-definite programming~\cite{fei2017exponential} algorithm, spectral clustering methods \cite{chin2015stochastic,rohe2011spectral,rohe2011spectral,baek2021review} possibly combined with iterative correction procedures~\cite{yun2014accurate,yun2016optimal,gao2017achieving}, or other iterative approaches~\cite{abbe2015community}.

Given the importance of graph clustering problems, there is still a need for faster and more efficient algorithms. Motivated by the Lloyd algorithm for the $k$-means problem \cite{lloyd1982least, du2006convergence, lu2016statistical, ahmed2020kmeans}, we introduce in this work a new simple and iterative algorithm for the label recovery and the parameter inference for the SBM problem for directed graphs. This algorithm is model-free in the sense that the functions to be minimized do not depend on the distribution of the edges. It is much faster than SDP or VEM type algorithms and empirically performs favorably in comparison the state of the art. Last but not least, we obtain consistency results for these algorithms. Therefore, we believe that this algorithm could be used as a building block in more complicated SBM models for fast and efficient label recovery.

\textbf{Outline}. In Section~\ref{sec:Lloyd}, we introduce our new algorithm and show how it follows naturally from rewriting of the log-likelihood as a $k$-means type problem. The consistency properties of this algorithm are presented in Section~\ref{sec:LloydSBM}. In Section~\ref{sec:Num_expe}, we compare the numerical performance of this algorithm with that of other commonly used algorithms, in terms of computation time and proportion of labels recovered, on simulated data. An application to a real dataset is provided in Section~\ref{sec:realdata}.

\section{Lloyd-type algorithms for the SBM problem}\label{sec:Lloyd}

In this section, we introduce our new algorithm for the labels recovery from a SBM. In Section~\ref{sec:SMBtoLloyd}, we show that the likelihood maximization problem is closed to a $k$-means problem in the case of the Bernoulli SBM. We also propose a first algorithm directly inspired by the Lloyd algorithm. In Section~\ref{sec:LloydSBM}, we propose the main algorithm of this article which is adapted to a wider class of SBMs, and we give consistency results on the associated estimator. The proof of these results is given in Appendix~\ref{app:prfthm}. \\

\subsection{Bernoulli SBM likelihood rewritten as a $k-$means problem}\label{sec:SMBtoLloyd}

Let $N\in\N^\star:=\N\backslash\{0\}$ be the number of nodes and $K\in\N^\star$ the number of labels. For all $i\in\intint{1}{N}:=\{1,\dots,N\}$, let $z_i\in\intint{1}{K}$ be the label of node $i$ and let $z=(z_i)_{i\in\intint{1}{N}}$ be the vector of labels. Let $P=(P_{pq})_{p,q\in\intint{1}{K}}\in[0,1]^{K\times K}$ be the probability matrix. We assume that the SBM $X=(X_{ij})_{i,j\in\intint{1}{N}}$ has $i.i.d.$ coordinates conditionally distributed to the labels as follows 
$$
\P\left(X_{ij}=1\right)=P_{z_iz_j},\quad\P\left(X_{ij}=0\right)=1-P_{z_iz_j}.
$$
It follows that the log-likelihood of this model is given by
\begin{equation}\label{eq:llh_sbm}
    \ell\left(X\middle|z,P\right)=\sum_{i,j=1}^N  \left(X_{ij} \log(P_{z_iz_j})+(1-X_{ij})\log(1-P_{z_iz_j}) \right),
\end{equation}
where we used the convention $0\times\log(0)=0$. Note that, unlike the VEM framework, the vector of labels is considered as a parameter. Let us give an alternative form for \eqref{eq:llh_sbm}. For all $z\in\intint{1}{K}^N$, let $I_p(z):=\{i\in\intint{1}{N}\ |\ z_i=p\}$ be the class of nodes with label $p\in\intint{1}{K}$. Let $N_p(z):=\card(I_p(z))$ be the number of nodes with label $p$. Let $\hmu_{iq}(z)$, $\hnu_{pj}(z)$ and $\hP_{pq}(z)$ be respectively the average number of edges from node $i$ to nodes of label $q$, from nodes of label $p$ to node $j$ and from nodes of label $p$ to nodes of label $q$. These quantities are given by

$$
\begin{aligned}
    &\hmu_{iq}(z)=\frac{1}{N_q(z)}\sum_{j\in I_q(z)}X_{ij} \quad \text{if}\ N_q(z)\ne 0,\qquad \hmu_{iq}(z)=0 \quad \text{otherwise}, \\
    &\hnu_{pj}(z)=\frac{1}{N_p(z)}\sum_{i\in I_p(z)}X_{ij} \quad \text{if}\ N_p(z)\ne 0,\qquad \hnu_{pj}(z)=0 \quad \text{otherwise}, \\
    &\hP_{pq}(z)=\frac{1}{N_p(z)N_q(z)}\sum_{i\in I_p(z)}\sum_{j\in I_q(z)}X_{ij} \quad \text{if}\ N_p(z)N_q(z)\ne 0,\qquad \hP_{pq}(z)=0 \quad \text{otherwise}.
\end{aligned}
$$

Note that if $N_p(z)N_q(z)\ne 0$, we have
\begin{equation}\label{eq:meanmu}
    \hP_{pq}(z)=\frac{1}{N_p(z)}\sum_{i\in I_p(z)}\hmu_{iq}(z)=\frac{1}{N_q(z)}\sum_{j\in I_q(z)}\hnu_{pj}(z).
\end{equation}

With this notation, the log-likelihood \eqref{eq:llh_sbm} can be written as
\begin{equation}\label{eq:llh_sbm_km}
    \ell\left(X\middle|z,P\right)=\sum_{p=1}^K \sum_{i\in I_p(z)}S_{ip}(z,P),
\end{equation}
where 
$$
S_{ip}(z,P)=  \overrightarrow{S}\!_{ip}(z,P):=  \sum_{q=1}^KN_q(z)\left(\hmu_{iq}(z)\log(P_{pq})+(1-\hmu_{iq}(z))\log(1-P_{pq})\right),
$$
or
$$
S_{ip}(z,P)=  \overleftarrow{S}\!_{ip}(z,P):= \sum_{q=1}^KN_q(z)\left(\hnu_{qi}(z)\log(P_{qp})+(1-\hnu_{qi}(z))\log(1-P_{qp})\right),
$$
or else
$$
S_{ip}(z,P)= \overline{S}_{ip}(z,P):= \frac{\overrightarrow{S}\!_{ip}(z,P) + \overleftarrow{S}\!_{ip}(z,P)}{2}.
$$
To maximize the likelihood $\ell\left(X\middle|z,P\right)$, we therefore want to maximize each term of the sum \eqref{eq:llh_sbm_km}. However, even though the three functions $\overrightarrow{S}$, $\overleftarrow{S}$, and $\overline{S}$ lead to the same expression of $\ell\left(X\middle|z,P\right)$, their use in an optimization scheme leads to different numerical performances. For instance, the choice $S=\overrightarrow{S}$ is more likely to fall into local minima since they only considers out and in edges respectively. It is therefore unable to identify classes if the $P$ matrix has equal rows. The problem is similar for $ S=\overleftarrow{S}$. The choice $S=\overline{S}$ counteracts these problems and produces great numerical results (see Section~\ref{sec:Num_expe}). We will therefore use $\overline{S}$ as $S$ function. \\

\subsection{The Lloyd-MLE-SBM algorithm}

Maximizing \eqref{eq:llh_sbm_km} is closed to a $k$-means problem. Recall that, given $\{x_i\}_{i\in\intint{1}{N}}$ a family of vector in $\R^d$ and $K\in\N^\star$, the $k$-means problem is to find a partition $(I_1,\dots,I_K)$ of $\intint{1}{N}$ minimizing the quantity:
\begin{equation}\label{eq:the_loss_kmeans}
    \sum_{p=1}^K\sum_{i\in I_p}\norm{x_i-\overline{x}_p}_2,
\end{equation}
where for all $p\in\intint{1}{K},\ \overline{x}_p = \sum_{i\in I_p}x_i/N_p$ and $N_p=\card(I_p)$. We note that setting $I_p = \{i\in\intint{1}{N}\ |\ z_i=p\}$ and $S_{ip}(z)=-\norm{x_i-\overline{x}_p(z)}_2$ where $\overline{x}_p = \sum_{i\in I_p(z)}x_i/N_p(z)$, Equation~\eqref{eq:the_loss_kmeans} can be written as \eqref{eq:llh_sbm_km}. Even if this problem is an NP-hard problem \cite{mahajan2012planar} as soon as $K\geq 2$, the Lloyd algorithm is an efficient iterative method that computes a local minimum. Inspired by this algorithm, we introduce Algorithm~\ref{alg:LloydMLE}, called hereafter Lloyd-MLE-SBM algorithm, for the SBM problem. \\

\begin{algorithm}[!ht]
\caption{The Lloyd-MLE-SBM algorithm }

\label{alg:LloydMLE}
\begin{algorithmic}[1]
\Require $z^{(0)}\in\intint{1}{K}^N$
\While {$z^{(n)}\ne z^{(n-1)}$ up to a permutation}
 \State $\hP^{(n)}\gets \hP(z^{(n)})$
 \For {$i\in \intint{1}{n}$}
    \State $\ z^{(n+1)}_i\gets \argmax_{p\in\intint{1}{N}} \overline{S}_{ip}(z^{(n)}, \hP^{(n)})$
 \EndFor
\EndWhile
\end{algorithmic}
\end{algorithm}

If Algorithm~\ref{alg:LloydMLE} converges to some $(\hz,\hP)$ then it satisfies the following Nash equilibrium type property: for every $z\in \intint{1}{K}^N$ such that there exists $i$ such that $\hz_j= z_j$ for all $j \neq i$, we have
$$
\ell(X|\hz,\hP) \, \geq \, \ell(X|z,\hP).
$$

\begin{rem}
Let us quickly draw a parallel between Algorithm~\ref{alg:LloydMLE} and game theory. The quantity $\overline{S}_{ip}(z^{(n)}, \hP^{(n)})$ corresponds to the payoff for an individual $i$ if he chooses to belong to class $z_i$ and the others are in class $z$. The iterations of Algorithm~\ref{alg:LloydMLE} then correspond to a repeated game in which each individual unilaterally chooses to change class according to the information it has: the classes of the other individuals at the previous instant. However, we want to optimize the sum of the payoffs, which in game theory is known as the social optimum. Unfortunately, it is well known in game theory that individual optimization without dialogue generally does not converge to the social optimum, as illustrated by the prisoner's dilemma. \\
\end{rem}

\subsection{The Lloyd-SBM algorithm}\label{sec:LloydSBM}

Instead of $\overleftarrow{S}\!_{ip}$, $\overrightarrow{S}\!_{ip}$ or, $\overline{S}_{ip}$ several
choices are possible for Algorithm~\ref{alg:LloydMLE}. Furthermore, to generalize Algorithm~\ref{alg:LloydMLE} to variants of SBM where $X_{ij}$ belongs to more general sets, we need to adapt the definition of $S_{ip}$. To that end, we observe that the function $\overline{S}\!_{ip}$ reaches its maximum when $(\hP(z)_{pq})_{q\in \intint{1}{K}}= (\hmu_{iq} (z))_{q\in \intint{1}{K}}$ and $(\hP(z)_{qp})_{q\in \intint{1}{K}}= (\hnu_{qi} (z))_{q\in \intint{1}{K}}$. Consequently, maximizing $\overline{S}\!_{ip}(z)$ is heuristically similar to both minimizing the distance between vectors $(\hP_{pq}(z))_{q\in \intint{1}{K}})$ and $(\hmu(z)_{iq})_{q\in \intint{1}{K}}$ and the one between $(\hP_{qp}(z))_{q\in \intint{1}{K}})$ and $(\hnu(z)_{qi})_{q\in \intint{1}{K}}$. Based on this observation, we introduce the main algorithm of this article below, called herefater Lloyd-SBM. \\

Let $\hpi_i(z)=(\hpi_{iq}(z))_{q\in\intint{1}{2K}}\in \R^{2K}$ be defined by $\hpi_{iq}(z)=\hmu_{iq}(z)$ if $q\in\intint{1}{K}$ and $\hpi_{iq}(z)=\hnu_{(q-K)i}(z)$ if $q\in\intint{K+1}{2K}$. Let $\hPi_p(z)=(\hPi_{pq}(z))_{q\in\intint{1}{2K}}\in \R^{2K}$ be defined by $\hPi_{pq}(z)=\hP_{pq}(z)$ if $q\in\intint{1}{K}$ and $\hPi_{pq}(z)=\hP_{(q-K)p}(z)$ if $q\in\intint{K+1}{2K}$. Here, $\hmu, \hnu$ and $\hP$ keep the same definition as in Section~\ref{sec:SMBtoLloyd}. Let finally $d$ be a distance on $\R^{2K}$. This leads to a heuristic variant of Algorithm~\ref{alg:LloydMLE} described in Algorithm~\ref{alg:LloydH}, which performs numerically well; see Section~\ref{sec:Num_expe} for details. \\

\begin{algorithm}
\caption{The Lloyd-SBM algorithm}
\label{alg:LloydH}
\begin{algorithmic}[1]
\Require {$z^{(0)}\in\intint{1}{K}^N$, a distance $d$}
\While {$z^{(n)}\ne z^{(n-1)}$ up to a permutation}
 \State $\hpi^{(n)}\gets \hpi\left(z^{(n)}\right)$ and $\hPi^{(n)}\gets \hPi\left(z^{(n)}\right)$
\For {$i\in \intint{1}{n}$}
    \State $\ z^{(n+1)}_i\gets \argmin_{p\in\intint{1}{N}}$\, $d\left(\hpi_i^{(n)},\hPi_p^{(n)} \right)$
 \EndFor
\EndWhile
\end{algorithmic}
\end{algorithm}

\subsection{Consistency of the related estimator}

The proof of the consistency of the estimator provided by Algorithm~\ref{alg:LloydH} is supported by the fact that two nodes with the same label will, in expectation, develop the same number of edges with nodes of another given label. More precisely, $(\hmu_{iq} (z^\star))_{i \in I_p(z^\star)}$ forms a sequence of $i.i.d$ random variables. Moreover, by the law of large numbers, $\hmu_{iq}(z)$ is close to its expectation for a large $N$. It follows that if $z=z^\star$ and $z_i=z_j$ then $\hmu_{iq}(z)$ is close to $\hmu_{jq}(z)$ and therefore to $\hP_{z_iq}(z)$ by \eqref{eq:meanmu}. We conclude that heuristically, $z^*$ is the vector of labels that minimizes the distance between $\hmu_{iq}(z)$ and their averages. This leads to the introduction of the estimator $\hz^N=\argmin\hL_N$ where

\begin{equation}\label{eq:Lfunction}
    \hL_N : z \mapsto \frac{1}{N} \sum_{p=1}^K\sum_{i\in I_p(z)} d\left(\hpi_i(z),\hPi_p(z)\right)=\frac{1}{N} \sum_{i=1}^N d\left(\hpi_i(z),\hPi_{z_i}(z)\right).
\end{equation}

Algorithm~\ref{alg:LloydH} is a minimization scheme of $\hL_N$. In the following, we will assume that $z^\star_i$ is the true label of node $i$. We will also assume that for all $i,j\in\intint{1}{N}$, $X_{ij}$ is an integrable $\R$-valued random variable such that $(X_{ij})_{i,j\in\intint{1}{N}}$ are independent given $z$ and $(X_{ij})_{i\in I_p(z^\star),j\in I_q(z^\star)}$ are identically distributed. Let $(P^\star_{pq})_{p,q\in\intint{1}{K}}$ be the expectations of these random variables, $i.e.$
$$
\forall i\in I_p(z^\star),\ j\in I_q(z^\star),\quad P^\star_{pq}=\E(X_{ij}). 
$$

More precisely, for any $N\in \N^*$, we consider a sequence of SBM $X=(X^N)_{N\geq 1}$ with parameter $z^\star=(z^{\star,N})_{N\geq 1}$ and a fixed matrix $(P^\star_{pq})_{p,q\in \intint{1}{K}}$. To prove consistency, we consider an identifiable framework and then consider the following assumption, coming from \cite[Assumption 1]{celisse2012consistency}.

\begin{ass}\label{ass:identifiability}
For all $p_1,p_2\in\intint{1}{K}$ such that $p_1\ne p_2$, there exists $q\in\intint{1}{K}$ such that $P^\star_{p_1 q}\neq P^\star_{p_2 q}$ or $P^\star_{q p_1}\neq P^\star_{q p_2}$, which can be rewritten as 
\begin{equation}\label{eq:identifiable_2}
\delta^\star:=\min_{p_1 \neq p_2}\max_{q}\left(\abs{P^\star_{p_1q}-P^\star_{p_2q}}+\abs{P^\star_{qp_1}-P^\star_{qp_2}}\right)>0.
\end{equation}
\end{ass}

Assumption~\ref{ass:identifiability} is a minimal identifiability condition. If this condition is not fulfilled, the label vector $z^\star$ cannot be estimated. Furthermore, as in \cite[Assumption 4]{celisse2012consistency}, we will assume that no class is asymptotically drained. \\

\begin{ass}\label{ass:notdrained}
There is $\gamma \in (0,1/K]$ such that for all $\gamma'<\gamma$ and $N$ large enough, 
$$
z^{\star,N}\in\Z^N_{\gamma'}:=\left\{z\in\intint{1}{K}^N\ \middle|\ \forall p\in\intint{1}{K}, N_p(z^\star)/N\geq\gamma'\right\}.
$$
\end{ass}
In the following, we introduce $\Z_\gamma$ the set of all sequences $z=(z^N)_{N\geq1}$ which satisfy Assumption~\ref{ass:notdrained} for some $\gamma>0$. \\

We are not in the classical framework of proving the consistency of estimators, since both the parameter of interest $z^\star$ and the state space $\intint{1}{K}^N$ depend on $N$. Moreover the label identification is not unique since we can do a permutation of the classes. To overcome these problems of definition, caused in part by the growing dimension of space, we will consider the function
\begin{equation}\label{eq:defDelta}
    \Delta = \Delta_N : (z,z') \mapsto \frac{K}{N^2(K-1)} \sum_{i,j=1}^N \1_{z_i=z_j}\1_{z'_i\ne z'_j}
\end{equation}
Number $\Delta(z,z')$ counts how many pair of nodes that belong to the same class for $z$ that are not in the same class for $z'$. It is not a pseudo-metric since $\Delta(z,z')=0$ does not imply $z=z'$ up to a permutation. However, if $\Delta(z,z')=0$ then there exists $\tau:\intint{1}{K}\to\intint{1}{K}$ which is not necessarily a permutation, such that $z'_i=\tau(z_i)$. But, if further $z,z'\in\Z^N_\gamma$ then $\tau$ is necessarily a permutation. Hereafter, we will say that our estimator $\hz$ is strongly consistent if $\Delta(\hz^N)=\Delta(z^{\star,N},\hz^N)$ converges almost surly to $0$ when $N$ goes to infinity. \\

Our consistency results is then provided by the following theorem proved in the particular case where $X_{ij} \in [0,1]$ and $d$ is the $\ell^1$-norm as a proof of concept. This is not a strong restriction since in finite dimension, all distances induced from norms are bounded by this norm.

\begin{thm}[Strong consistency of $\hz$]\label{thm:consist}
Suppose $X_{ij} \in [0,1]$, $d$ is the $\ell^1$-norm, and that Assumption~\ref{ass:identifiability} and Assumption~\ref{ass:notdrained} hold. Let $\hz^N\in \argmin \hL_N$ and suppose further that there is $\gamma_0\in(0,1/K]$ such that $\hz\in\Z_{\gamma_0}$ and $\delta_0>0$ such that for $N$ large enough,
\begin{equation}\label{eq:meanindentif}
\min_{p_1\neq p_2} \max_{q} \left| \bP_{p_1 q}(\hz^N) - \bP_{p_2 q}(\hz^N) \right| + \left| \bP_{q p_1}(\hz^N)- \bP_{q p_2}(\hz^N) \right|\geq\delta_0,
\end{equation}
where $\bP_{pq}(z):=\E\left[\hP_{pq}(z)\right]$. Then $\Delta(\hz^N)$ and $\sup_{z\in\Z_{\gamma_0}}\norm{\hP(z)-\bP(z)}$ converge almost surely to $0$. More precisely, for any $r<1/3$ and $N$ large enough, there exists $C>0$ such that $\Delta(\hz^N)\leq C N^{-r}$ with probability at least $1-e^{-N}$.
\end{thm}

The proof is detailed in Appendix~\ref{app:prfthm}. However, note that  our algorithm may not lead to the global minimum of $\hL_N$ in $\Z^N_{\gamma_0}$ or a minimum verifying \eqref{eq:meanindentif}. However, Theorem~\ref{thm:consist} ensures that, if
$$
\forall p \in \intint{1}{K}, \quad \frac{N_p(\hz)}{N}\gg 0, \quad \text{ and } \quad \forall p_1,p_2 \in \intint{1}{K}, \quad \max_{q\in\intint{1}{K}}| \hP_{p_1 q}(z) - \hP_{p_2 q}(z)| + |\hP_{q p_1}(z)- \hP_{q p_1}(z)| \gg 0,
$$
then we may be confident in our estimation. Note also that Assumption~\eqref{eq:meanindentif}, which states that our estimator verifies the identifiability assumption we state for the parameter to be estimated, is quite mandatory to prove consistence since $\hL_N$ may admit others minimizer; see Remark~\ref{rem:K=2}.

\section{Numerical Study}\label{sec:Num_expe}

This section summarizes numerical experiments to evaluate the algorithmic performance of algorithms presented in Section~\ref{sec:Lloyd}. A more complete analysis is presented in Appendices~\ref{app:compLloydHuber} and \ref{app:othermethods}. \\

In Section~\ref{sec:protocol} below, we present the general protocol we use to compare these algorithms. In Section~\ref{sec:LloydHuber}, we compare variants of Algorithm~\ref{alg:LloydH} for different choices of distance $d$. Our study reveals that the choice $d(x,y)=\norm{x-y}_1$ leads to slightly better performance than the others, especially when initialized with spectral clustering. In Section~\ref{sec:otheralgorithms}, we compare Algorithm~\ref{alg:LloydH} with three other commonly used inference methods: spectral clustering, gradient descent and variational EM, which are presented in Appendix~\ref{app:Inf_methods}. 

\subsection{General experimental protocol}\label{sec:protocol}

We set the number of clusters at $K=3$ and we consider two forms of probability matrices:
$$
P_{sym}(a,b) = \left(\begin{array}{ccc}
    a & b & b \\
    b & a & b \\
    b & b & a 
\end{array}\right)\quad\text{and}\quad P_{asym}(a,b) = \left(\begin{array}{ccc}
    a & b & b' \\
    b' & a & b \\
    b & b' & a 
\end{array}\right),
$$
where $a,b\in[0,1]$ and $b'=b+(a-b)/a$ is chosen to satisfy the following symmetry-weighted equality $(a-b)/a = (b'-a)/(1-a)$. These two matrices satisfy Assumption~\ref{ass:identifiability} if and only if $a \ne b$. It follows that the smaller $\abs{a-b}$ is, the less identifiable the classes of nodes are. If the parameter $a$ is fixed, the identifiability of the model then depends on $b$, which will be referred to as the identifiability parameter. All our numerical experiments lead to the same conclusions whether using $P_{sym}$ or $P_{asym}$. Consequently, for the sake of concision, we will only detailed the case $P=P_{asym}$ in the following. The results for $P_{sym}$ are only presented in table form (see Tables~\ref{tab:1}, \ref{tab:2}, \ref{tab:3} and \ref{tab:4} in Appendix~\ref{app:othermethods}). \\

Since the Lloyd algorithm generally produces clusters of the same size, known as the uniform effect \cite{wu2012uniform}, we will test the algorithm's performance with heterogeneous cluster sizes. More precisely, for every node $i$, the label $z^\star_i$ will be drawn independently according to the distribution
\begin{equation}\label{eq:defalpha}
    \alpha(h) = \left(\dfrac{1-h}{3},\dfrac{1}{3},\dfrac{1+h}{3}\right),
\end{equation}
where $h\in[0,1]$ will be referred to as the heterogeneity parameter. \\

For iterative methods, we will evaluate the performance of algorithms with respect to the initialization $z^{(0)}$. On the one hand, initialization can be done with a noisy version of $z^\star$ where each label in $z^\star$ is replaced with probability $\omega$ by a label drawn uniformly from $\intint{1}{K}$. Note that if $N$ is small, then it is possible that the corresponding initial vector of labels does not contain all the different labels. To avoid this problem, this simulation procedure is repeated until $z^{(0)}$ contains all the labels. We will also compare the performance of each iterative method according to the parameter $\omega$ to evaluate their sensitivity to the distance between $z^{(0)}$ and $z^\star$. On the other hand, we will compare the iterative algorithms when initialized by the spectral method. \\

\begin{table}[!ht]
    \centering
    \begin{tabular}{|c|c|c|}
        \hline
        Name & Notation & Description \\
        \hline
        Identifiability parameter & $b$ & Distance between $P$ and the non identifiable case\\
        \hline
        Heterogeneity parameter & $h$ & Size heterogeneity of classes of $z^\star$\\
        \hline
        Noise parameter & $\omega$ & Proportional to the distance between $z^{(0)}$ and the $z^\star$\\
        \hline
    \end{tabular}
    \caption{Summary of variable parameters in numerical experiments.}
    \label{tab:sumparam}
\end{table}

\subsection{Results of numerical experiments}

To evaluate whether an estimated label $\hz$ is close to the true label $z^\star$, we will use $\Gamma(\hz)=\Gamma(z^\star,\hz)$ defined by 

\begin{equation}\label{eq:defGamma}
    \Gamma = \Gamma_N : (z,z') \mapsto \frac{K}{2N^2(K-1)} \sum_{i,j=1}^N \abs{\1_{z_i=z_j}-\1_{z'_i\ne z'_j}}.
\end{equation}

Note that $\Gamma$ is a pseudo-distance on the label space and, in particular, $\Gamma(z,z')=0$ if and only if $z=z'$ up to a permutation. In addition, we will study the speed performance through the computation time measured in seconds. \\

The simulations\footnote{https://github.com/acotil/LloydSBM} were performed using the Julia language on a computer with a Changer 2.1 GHz i7 (4 cores) processor and 32 GB of memory. For all the iterative methods, we will fix the maximum number of iterations to $100$ and the maximum time of computation to $10\ \text{sec}$.

\subsubsection{Performance of Algorithm~\ref{alg:LloydH}}\label{sec:LloydHuber}

We test Algorithm~\ref{alg:LloydH} with the $\ell_1$-norm, the $\ell_2$-norm and an interpolation between these two distances, corresponding to the Huber loss. For these first two cases, Algorithm~\ref{alg:LloydH} will be referred to as the "LloydL1" and "LloydL2" methods respectively. The Huber loss, introduced in \cite{huber1992robust} in the context of robust parameter estimation, with parameter $r>0$ between two vectors $x,y\in\R^N$ is given by
\begin{equation}\label{eq:Huber}
    \mathrm{Huber}_r(x,y) = \frac{1}{2}\sum_{i=1}^N(x_i-y_i)^2\1_{\abs{x_i-y_i}\leq r}+\left(2r\abs{x_i-y_i}-r^2\right)\1_{\abs{x_i-y_i}>r}.
\end{equation}
Again, we will refer to the corresponding algorithm as "LloydHuber". After a large number of tests, presented in Appendix~\ref{app:compLloydHuber}, we conclude that the "Lloyd" methods are broadly similar in terms of $\Gamma$ and computation time, for all the ranges of tested parameter values. The results of these tests are summarized in Tables~\ref{tab:1} and \ref{tab:2}. We will use the "LloydL1" method to compare the "Lloyd" methods with the state-of-the-art algorithms. 

\subsubsection{Comparison between performance of Algorithm~\ref{alg:LloydH} and state-of-the-art methods}\label{sec:otheralgorithms}

The methods compared in this paper are divided into 4 classes: The first corresponds to the Lloyd-type methods introduced in this article. The second class correspond to the gradient descent type methods. The third class are variational EM methods. Finally, the last class of methods is those of the spectral clustering type. A detailed presentation of these algorithms is given in Appendix~\ref{app:Inf_methods}. \\

The comparison between the "LloydL1", "LloydMLE" (corresponding to Algorithm~\ref{alg:LloydMLE}), "GD" (for gradient descent) and "VEM" methods shows first that the “GD” and "VEM" methods are much slower than the "LloydL1" and "LloydMLE" methods. In addition, the use of each method after spectral clustering shows a significant increase in performance compared to spectral clustering alone. The “LloydL1” method has proven to offer an optimal balance between estimation quality in terms of $\Gamma$ and computation time. In comparison with alternative methods, it has been demonstrated to be particularly effective when the number of nodes is limited or when classes are not readily identifiable, while exhibiting exceptional speed, even for graphs of considerable size. The results are summarized in Tables~\ref{tab:3} and \ref{tab:4}. \\

\section{Applications to social role inference in animals with real-world dataset}
\label{sec:realdata}

In this section, we present an application of Algorithm~\ref{alg:LloydH} on the dataset presented in \cite{leroux2021growth}. It is recorded by an experimental prototype, called Walk-over-Weighting (WoW), designed for automatic weighing of ewes. This device, installed in an outdoor setting, corresponds to a gate through which the ewes must pass if they wish to reach their feed trough, and which records the time of passage, the identity of the individual and its weight. It was originally developed to study ewes growth in real time. \\

\begin{figure}[!ht]
    \centering
    \includegraphics[scale=0.5]{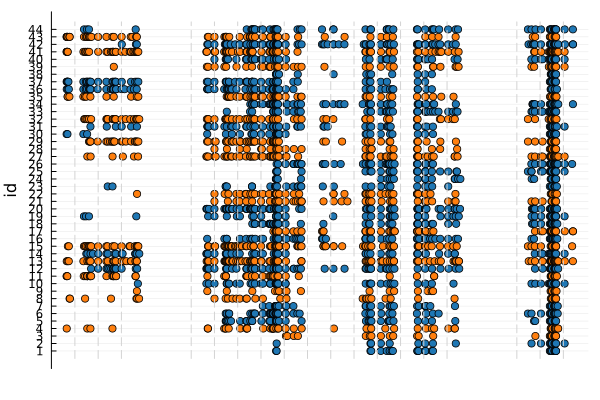}
    \caption{Time each ewe passed through the WoW device. Dotted lines represent the start of each day at midnight. Colors represent the label estimated by Algorithm~\ref{alg:LloydH} for $K=2$.}
    \label{fig:passage_time_2}
\end{figure}

\begin{figure}[!ht]
    \centering
    \includegraphics[scale=0.5]{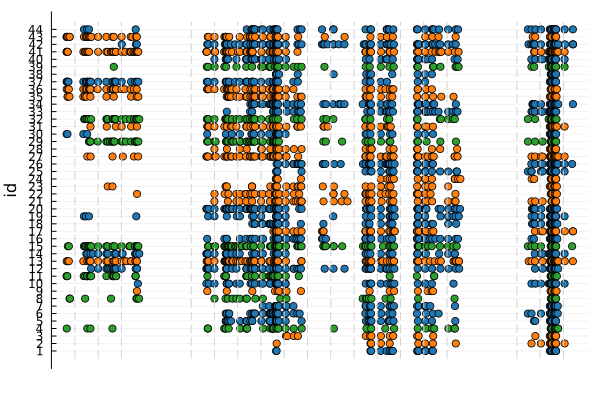}
    \caption{Time each ewe passed through the WoW device. Dotted lines represent the start of each day at midnight. Colors represent the label estimated by Algorithm~\ref{alg:LloydH} for $K=3$.}
    \label{fig:passage_time_3}
\end{figure}

For this study, we will focus only on the times at which the ewes pass through the device. More precisely, we will study the passage times of a herd of $44$ ewes (numbered $1$ to $44$), collected with the WoW device over one month (mid-May to mid-June 2017) in France at the experimental farm of the Domaine du Merle (see Figure~\ref{fig:passage_time_2} and \ref{fig:passage_time_3}). \\

We aim to classify the animals according to whether they crossed the device rather in a group or rather in isolation. To do this, we will first use this dataset to create a weighted graph where each node $i\in\intint{1}{44}$ is associated with an individual and where the weight of the edge $(i,j)$ is proportional to the number of times individuals $i$ and $j$ crossed the device at one-minute intervals. More precisely, we define 
$$
A_{ij} = \frac{1}{2}\card\left\{(s,t)\in T_i\times T_j\ \middle|\ \abs{s-t}\leq 1\right\},
$$
where $T_i$ is the set of times in minutes that individual $i$ has passed through the device. To take into account the effect of the total number of times an individual has passed through the device, we assume that the adjacency matrix $X$ of the graph on which we will apply our algorithm is given by
$$
X_{ij} = \frac{A_{ij}}{\sqrt{n_i}\sqrt{n_j}},
$$
where $n_i$ is the total number of times that individual $i$ has passed through the device. Since $n_i$ is related to the degree of node $i$, this choice of normalization of matrix $A$ was made for the sake of homogeneity (as in \cite{chaudhuri2012spectral}). 
Note that in this case, it seems difficult to model $X_{ij}$ as a random variable whose distribution is standard or easy to evaluate, which makes the use of Algorithm~\ref{alg:LloydH} particularly relevant since it is model-free. \\

\begin{figure}[!ht]
    \centering
    \includegraphics[scale=0.5]{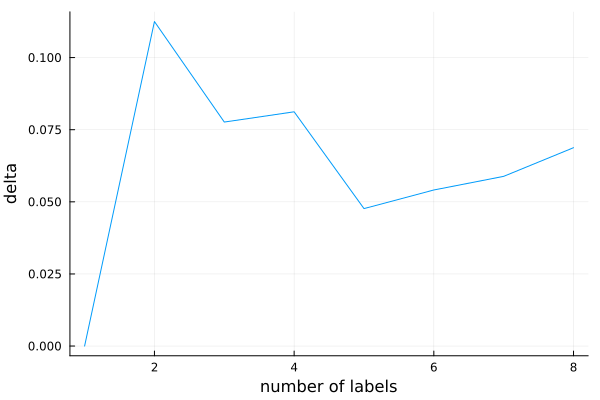}
    \caption{Evolution of $\hdelta$ according to the desired number of class $K$.}
    \label{fig:delta}
\end{figure}

For the inference, we have chosen the $\ell_1$-norm for distance $d$. 
To choose the number of classes $K$, we intended, as usual, to use the ICL criterion. A first problem is that, since the distribution used on $X_{i,j}$ is not standard, we do not have access to an explicit formula for the likelihood. To overcome this problem, we numerically implemented an approximate likelihood using a beta distribution with the same moments. However, this did not give us satisfactory results for choosing the number of classes $K$. One reason for this is surely that the ICL criterion itself is a suitable approximation for larger-scale networks. For these reasons, we have therefore based our selection of the number of classes on the constant $\hdelta$ defined by
$$
\hdelta = \min_{p_1\neq p_2} \max_{q} \left| \hP_{p_1 q}(\hz) - \hP_{p_2 q}(\hz) \right| + \left| \hP_{q p_1}(\hz)- \hP_{q p_2}(\hz) \right|.
$$
In light of our theoretical results, the constant $\hdelta$ is, in a sense, an estimator of a quantity measuring the identifiability of the system with $K$ classes. The idea is to choose a $K$ that gives rise to identifiable classes. This technique does not allow us to select an optimal $K$, but rather admissible $K$ values that give rise to identifiable classes. Figure~\ref{fig:delta} shows that $\hdelta$ is maximal for $K=2$. \\

In Figures~\ref{fig:passage_time_2} and \ref{fig:passage_time_3} each class is represented by a color. For the estimation, we start by perform Algorithm~\ref{alg:LloydH} where initialized with Spectral clustering. Then, we repeat the process of drawing uniformly at random $z^{(0)}$ and computing $\hz$ using Algorithm~\ref{alg:LloydH} for $300$ seconds. We select the estimate $\hz$ that minimizes \eqref{eq:Lfunction}. A total of $2546794$ estimations were perform for $K=2$ and $714613$ for $K=3$. For all tested $K$, we systematically observe that initialization by spectral clustering does not result in an estimate that maximizes $L$. The estimated expectation matrices $\hP=\hP_K$ for $K\in\{2,3\}$ are given by
$$
\hP_2=\left(\begin{array}{cc}
    0.16 & 0.11 \\
    0.11 & 0.09
\end{array}\right)\quad
\hP_3=\left(\begin{array}{ccc}
    0.17 & 0.13 & 0.08 \\
    0.13 & 0.11 & 0.09 \\
    0.08 & 0.09 & 0.08 
\end{array}\right).
$$
According to these results, we observe for $K=2$ that there is individuals who have often crossed the device together (the blue class) and a class of individuals who have crossed it rather individually (the orange class). for $K=3$, classes $2$ and $3$ (respectively orange and green in \ref{fig:passage_time_3}) seems two almost be sub-classes of the previous orange class.

\section{Discussion}

The aim of this work was to study the performance of a new label inference method for nodes clustering in graphs. This method is based on a rewriting of the likelihood of the model, which allows us to map the problem of label inference in an SBM to that of $k$-means type problem. Based on this mapping, we propose a new Lloyd-type heuristic algorithm. This algorithm turns out to be much faster than the algorithms commonly used in the literature for the ranges of parameter values we tested. It also proves to be consistent whatever the distribution of edge weights conditional on node labels. This is a particularly important property for the study of real datasets, since in most cases it is not possible to know this distribution explicitly. It should also be noted that it is certainly possible to obtain more detailed results on the consistency of the estimator and the local convergence of the algorithm. This will be the subject of future work. \\

Let us also note that our method is based on how $\hmu_{ip}, \hnu_{ip}$ remains constant over all nodes $i$ in the same class $p$. We choose the mean for these quantities and our algorithm will be particularly relevant when the means of $X_{ij}$ are different over all the classes (see Assumption~\ref{ass:identifiability} and Assumption~\ref{ass:notdrained}). We can nevertheless choose any other functional and do the same type of optimization. We are thinking in particular of taking other $M$-estimators, such as the maximum likelihood associated with the SBM model variant. \\

Finally, this algorithm is studied in a simple version that does not include automatic determination of the number of classes. Merge and split techniques (based on a ICL type criterion adapted to estimator $\hz$) \cite{Chiquet2024Rsbm} and a variant adapted to degree-corrected SBM \cite{karrer2011stochastic} can also be considered. \\

\textbf{Acknowledgement.} Part of this work was carried out during A. Cotil's PhD thesis in the MISTEA and SELMET units. This work was supported by the French National Research Agency under the Investments for the Future Program, referred as ANR-16-CONV-0004 and by the EU project TechCare, which received funding from the European Union’s Horizon 2020 Research and Innovation Programme under grant agreement No 862050..

\bibliographystyle{abbrv}
\bibliography{refs.bib}

\appendix

\section{Proof of Theorem~\ref{thm:consist}}\label{app:prfthm}

In the following, let us define $\bmu_{iq}(z)=\E\left[\hmu_{iq}(z)\right]$, $\bnu_{qi}(z)=\E\left[\hnu_{qi}(z)\right]$, $\bpi_{iq}(z)=\E\left[\hpi_{iq}(z)\right]$, $\bP_{pq}(z)=\E\left[\hP_{pq}(z)\right]$ and $\bPi_{pq}(z)=\E\left[\hPi_{pq}(z)\right]$. Let us also define 
\begin{equation*}
\bL_N (z) = \frac{1}{N} \sum_{p=1}^K\sum_{i\in I_p(z)} d\left(\bpi_i(z),\bPi_p(z)\right)= \frac{1}{N} \sum_{i=1}^N \sum_{q =1}^K \left( \left| \bmu_{iq} - \bP_{pq}(z) \right| + \left| \bnu_{qi} - \bP_{q p}(z) \right|\right),\\
\end{equation*}
and
\begin{equation*}
\bdelta (z) = \min_{p_1\neq p_2} \max_{q} \left| \bP_{p_1 q}(z) - \bP_{p_2 q}(z) \right| + \left| \bP_{q p_1}(z)- \bP_{q p_2}(z) \right|.
\end{equation*}

Let us fix $i,j \in I_p(z^\star)$. We then have for all $z\in\intint{1}{K}^N$ and $q \in \intint{1}{K}$, $\bmu_{iq}(z) = \bmu_{jq}(z)$, and $\bnu_{qi}(z) = \bnu_{qj}(z)$. Using this, triangular inequality and Equation~\eqref{eq:meanindentif} gives 
\begin{align*}
    \mathbf{1}_{\{z_i \neq z_j\} } \bdelta(z) 
    &\leq \max_{q}\left(\left| \bP_{z_i q}(z) - \bP_{z_j q}(z) \right| + \left| \bP_{q z_i}(z)- \bP_{q z_j}(z) \right|\right)\\
    &\leq \max_{q}\left(\left| \bmu_{iq}(z)- \bP_{z_i q}(z) \right| + \left| \bmu_{j q}(z) - \bP_{z_j q}(z) \right|\right. \\
  &+ \left.\left| \bnu_{q j}(z) - \bP_{q z_j}(z) \right| + \left| \bnu_{q i}(z) - \bP_{q z_i}(z) \right|\right).
\end{align*}

Forgetting some dependencies on $z$ for the sake of presentation, this leads to
\begin{align*}
 \Delta(z)\bdelta(z)&=\frac{K}{N^2(K-1)} \sum_{i=1}^N \sum_{j \in I_{z^\star_{i}} (z^\star)} \mathbf{1}_{z_{i} \neq z_{j}}\bdelta(z) \\
 &\leq\frac{K}{N^2(K-1)} \sum_{i=1}^N \sum_{j \in I_{z^\star_{i}}(z^\star)}\max_{q}\left(\left| \bmu_{iq}- \bP_{z_i q} \right| + \left| \bmu_{jq} - \bP_{z_jq} \right|+\left| \bnu_{qj} - \bP_{qz_j} \right| + \left| \bnu_{qi} - \bP_{qz_i} \right|\right) \\
 &\leq\frac{2K}{N^2(K-1)} \sum_{p=1}^K \sum_{i \in I_p(z^\star)}N_p(z^\star)\max_{q}\left(\left| \bmu_{iq}- \bP_{z_iq} \right|+\left| \bnu_{qi} - \bP_{q z_i} \right|\right) \\
 &\leq\frac{2K}{N(K-1)} \sum_{i=1}^N\sum_{q=1}^K\left(\left| \bmu_{iq}- \bP_{z_i q} \right|+\left| \bnu_{qi} - \bP_{q z_i} \right|\right) =\frac{2K}{(K-1)}\bL_N(z).
\end{align*}
Then, if $\bdelta(z)>0$, we have
\begin{align*}
\Delta(z) &\leq \frac{2K}{(K-1) \bdelta(z)} \ \bL(z). 
\end{align*}

We will now bound for $\max_{z\in \Z_{\gamma'}^N}|\hL_N(z)- \bL_N(z)|$ for $\gamma'\in(0,\gamma)$. For $z\in \Z_{\gamma'}^N$, triangular inequality ensures that

\begin{align}
&\abs{ \hL_N(z) -\bL_N(z)} \nonumber \\
 = &\abs{\frac{1}{N}\sum_{i=1}^N d(\hpi_i(z),\hPi_{z_i}(z))-\frac{1}{N}\sum_{i=1}^N d(\bpi_i(z),\bPi_{z_i}(z))}\nonumber \\
\leq &\frac{1}{N}\sum_{i=1}^N d(\hpi_i(z),\bpi_i(z))+\frac{1}{N}\sum_{p=1}^K N_p(z)d(\hPi_p(z),\bPi_p(z)) \\
=&A(z)+B(z)+C(z),
\end{align}
where
$$
A=\frac{1}{N}\sum_{i=1}^N\sum_{p=1}^K\abs{\hmu_{ip}-\bmu_{ip}},\quad B=\frac{1}{N}\sum_{i=1}^N\sum_{p=1}^K\abs{\hnu_{pi}-\bnu_{pi}}\quad\text{and}\quad C=\frac{1}{N}\sum_{p=1}^K\sum_{q=1}^K (N_p+N_q)\abs{\hP_{pq}-\bP_{pq}}
$$

Let us begin by bounding $A(z)$. To that end, let us fix $r>0$. Since $\hmu_{ip},\bmu_{ip}\in[0,1]$, we have 
$$
A(z)\leq KN^{-r}+ \frac{1}{N} \sum_{i=1}^N \sum_{p=1}^K \mathbf{1}_{|\hmu_{ip}(z)- \bmu_{ip}(z)|\geq N^{-r}}.
$$
Since $N_p(z)\geq \gamma' N$ for $z \in \Z_{\gamma'}^N$ and $N$ large enough, Hoeffding's inequality gives that, for any $i$ and any $p$, we have 
$$
\P\left[|\hmu_{ip}(z)- \bmu_{ip}(z)|> N^{-r}\right]\leq 2\exp(-2 N_p(z) N^{-2r} )\leq 2e^{-2\gamma' N^{1-2r} }. 
$$
Since the $(\hmu_{iq})_{i\in\intint{1}{N},q\in\intint{1}{K}}$ are independent random variables, it follows that $N(A(z)-KN^{-r})$ is stochastically dominated by a Binomial random variable with parameters $(NK, s)$ with $s=s_N= 2e^{-2\gamma' N^{1-2r}}$. We deduce from the Bennett inequality for Binomial random variable that 
$$
\P\left( N(A(z)-KN^{-r}) - KNs > t\right)\leq \exp(-NKs\times h\left(\frac{t}{NKs}\right)),
$$
which leads to 
$$
\P\left(A(z) > KN^{-r} + Ks + t\right)\leq \exp(-NKs\times h\left(\frac{t}{Ks}\right)),
$$
where $h: u \mapsto (1+u)\log(1+u)-u$. Using the fact that for all $x\geq0$, $h(u)\geq u\log(u)/2$, we have 
$$
\P\left(A(z) > KN^{-r} + Ks + t \right)\leq \exp(-\frac{Nt}{2}\log(\frac{t}{Ks})).
$$
Taking $t=2KN^{-ar}$, we have
$$
\P\left(A(z) > K(N^{-r} + s + 2N^{-ar})\right)\leq\exp(-2\gamma'KN^{2-2r-ar}+arK\log(N)N^{1-ar}).
$$
If there exists $a\geq1$ and $r<1/2$ (so that $s+2N^{-ar}=o(N^{-r})$) such that 
$$
\lim_{N\to+\infty}2\gamma'N^{2-2r-ar}-aK\log(N)N^{1-ar}-(1+\log(K))N=+\infty,
$$
then, for all $\varepsilon>0$, there exists $N_\varepsilon>0$ such that $\forall N\geq N_\varepsilon$ we have
$$
\P\left(A(z) > (1+\varepsilon)KN^{-r})\right)\leq \frac{K^{-N}e^{-N}}{3}.
$$
This is possible if $2-2r-ar>1$ and thus, together with $a\geq1$, if $r<1/3$. Using an union bound, it follows that if $r<1/3$, then for $N$ large enough 
$$
\P\left(\max_{z\in\Z_{\gamma'}^N}A(z) > 2KN^{-r}\right) \leq \frac{e^{-N}}{3}
$$
In the same way, we have for $N$ large enough if $r<1/3$, 
$$
\P\left(\max_{z\in\Z_{\gamma'}^N}B(z) > 2KN^{-r}\right)\leq \frac{e^{-N}}{3}.
$$
Let us turn to the last term. For a fixed $z$, we have 
\[
C(z)\leq 2K^2 \max_{p,q} \frac{|V_{pq}(z)|}{N_p(z) N_q(z)} \ , 
\] 
where $V_{pq}(z) = \sum_{i\in I_p(z)} \sum_{j\in I_q(z)} (X_{ij} - \E[X_{ij}])$ is a centered sum of independent bounded random variables. By Hoeffding's inequality and a union bound, if $r<1/2$, we have 
\[
\P[C(z)\geq 2K^2 N^{-r}]\leq 2K^2\max_{p,q}\exp( - 2 N_{p}(z)N_{q}(z)N^{-2r})\leq 2K^{2}\exp(-2 \gamma^{'2} N^{2(1-r)})\leq \frac{K^{-N}e^{-N}}{3} \ ,
\] 
for $N$ large enough. By a union bound over all such $z$ and together with previous results, we conclude that if $r<1/3$,
\[
    \max_{z\in\Z_{\gamma'}^N}|\hL_N(z)- \bL_N(z)|\leq 6K^2 N^{-r}\ , 
\]
with probability larger than $1-e^{-N}$. Since we have $z^\star\in\Z_\gamma$ and $\hz\in\Z_{\gamma_0}$, we have for all $0<\gamma'<\min(\gamma,\gamma_0)$ and $N$ large enough, 
$$
\bL_N(\hz^N)= \hL_N(\hz^N) - \bL_N(z^{\star,N}) + \bL_N(\hz^N) - \hL_N(\hz^N) \leq 2\max_{z\in\Z_{\gamma'}^N}|\hL_N(z)- \bL_N(z)|.
$$
Since for $N$ large enough, $\bdelta(\hz^N)\geq\delta_0$, we have
$$
\Delta(\hz^N)\leq \frac{24K^3N^{-r}}{\delta_0(K-1)},
$$
with probability larger than $1-e^{-N}$. It follows that for all $\varepsilon>0$ and for $N$ large enough,
$$
\P\left(\Delta(\hz^N)>\epsilon\right)\leq e^{-N},
$$
which leads to the almost sure convergence using the Borel–Cantelli lemma. The same result for $\sup_{z\in\Z_{\gamma_0}}\norm{\hP(z)-\bP(z)}$ directly comes from the control of $C(z)$ and a union bound. \\

\begin{rem}\label{rem:K=2}
Let us compute $\bL_N$ on the simpler example of the Bernouilli SBM. In this setting, $P^\star_{pq}$ represent both the expectation of $X_{ij}$ if $i\in I_p(z^\star)$ and $j\in I_q(z^\star)$ and the probability that it is equal to $1$. We also assume that $K=2$ and 
$$
P^\star=\left(\begin{array}{cc} a & b \\ b & a \end{array}\right).
$$
In the case $d(x,y)=\norm{x-y}_1$, we have
$$
\bL_N(z)=4\abs{a-b}\left(\sum_{p=1}^2\frac{N_{p1}(z,z^\star)N_{p2}(z,z^\star)}{N_p(z)}\right)\left(\sum_{p=1}^2\frac{\abs{N_{p1}(z,z^\star)-N_{p2}(z,z^\star)}}{N_p(z)}\right).
$$
In the one hand, if $z\in\Z_{\gamma_0}$ for $\gamma_0>0$, we have 
$$
\sum_{p=1}^2\frac{N_{p1}(z,z^\star)N_{p2}(z,z^\star)}{N_p(z)}=0 \quad \iff \quad z=z^\star\text{ up to a permutation}.
$$
which implies that the vectors of labels $z$ equal to $z\star$ up to a permutation belong to the minimizers of $\bL_N$. On the other hand, we also note that 
$$
\sum_{p=1}^2\frac{\abs{N_{p1}(z,z^\star)-N_{p2}(z,z^\star)}}{N_p(z)}=0 \quad \iff \quad \forall p\in\{1,2\},\ N_{p1}=N_{p2}=N_p(z^\star)/2.
$$
This implies that the vectors that divided the true classes into two equal parts also belongs to the minimizers of $\bL_N$. Note that these vectors are those that maximize the distance to $z^\star$ up to a permutation. However, we easily verify they also verify
$$
\forall p\in\{1,2\},\ \bP_{p1}=\bP_{p2},
$$
and thus, from the law of large numbers,
$$
\forall p\in\{1,2\},\ \hP_{p1}\simeq\hP_{p2},
$$
which allows to exclude them using Theorem~\ref{thm:consist} in the limit of large networks.
\end{rem}

\section{Description of clustering algorithms used for comparison}\label{app:Inf_methods}

\textbf{Spectral clustering type methods :} The general idea of the spectral clustering for the SBM is that if the adjacency matrix is closed to a block matrix then its eigenvectors are closed to block vectors. The algorithm we use is the one described in \cite{deng2024subsampling} with the regularization introduced in \cite{amini2013pseudo}. Note that spectral clustering is only possible if the adjacency matrix is symmetric (i.e. if the corresponding graph is undirected). We will therefore apply spectral clustering by replacing the notion of eigenvector by that of singular vector, as is done in \cite{deng2024subsampling}. In the following, this method will be reffered as the "Spectral" method. For the sake of completeness, the algorithm we implemented is described by Algorithm \ref{alg:Spectral}. We define $\lambda_{reg}\in[0,1]$ the regularization coefficient used in the algorithm. In the following, $\1_{N\times N}$ will denote the $N\times N$-matrix whose coefficients are all $1$. \\ 

\begin{algorithm}
\caption{The regularized Spectral algorithm}
\label{alg:Spectral}
\begin{algorithmic}[1]
\State $Y\gets (X')^t(X')$ where $X'\gets X+\lambda_{reg}\times\left(\sum_{i,j}X_{ij}/N^2\right)\times\1_{N\times N}$.
\State $L\gets (D)^{-1/2}Y(D)^{-1/2}$ where $D\gets \mathrm{diag}\left(\left(\sum_jY_{ij}\right)_{i\in\intint{1}{N}}\right)$.
\State $U\gets\left(u_1\ \dots\ u_K\right)\in\R^{N\times K}$ where $u_p$ is the eigenvector of $L$ corresponding to its $p$-th largest eigenvalue. 
\State Runs k-means with $K$ clusters on the rows of $U$, seen as $N$ vectors of $\R^K$.
\end{algorithmic}
\end{algorithm}

\textbf{Lloyd type methods :} These methods are generally described by Algorithm~\ref{alg:LloydH}. Each method is defined by the $d$ metric used. This metric may be the maximum likelihood metric (corresponding to Algorithm~\ref{alg:LloydMLE}), in which case it is referred to as the "LloydMLE" method. We will also consider the case where $d$ correspond to the distance related to the $\ell_1$-norm. This method will be refer to as "LloydL1". We have also considered other metrics similar to the $\ell_1$-norm ($\ell_2$-norm and Huber loss for different parameter values). For each of these methods, we can either initialize it with a noisy version of the true labeling (the one obtained by simulation) or with the one obtained by spectral clustering. If the spectral clustering is used for the initialization, we will refer those methods as "Spectral+LloydMLE", "Spectral+LloydL1". \\

\textbf{Gradient descent type methods :} This type of method involves maximizing the likelihood locally with respect to each individual's class. Based on the work of Karrer and Newman \cite{karrer2011stochastic}, it was used in \cite{serrano2024community} as an alternative to the EM algorithm in order to compare it with the performance of an exact recovery algorithm. We call this method "gradient descent" since the principle of locally optimizing an objective function to find the global optimum is at the heart of standard gradient descent algorithms. Defining $\Hat{l}(z)$ as
$$
\Hat{l}(z) := l\left(X\middle|z,\Hat{P}(z)\right)=\sum_{i,j=1}^N \left(X_{ij} \log(\Hat{P}_{z_iz_j}(z))+(1-X_{ij})\log(1-\Hat{P}_{z_iz_j}(z)) \right),
$$
for a given label $\Hat{z}^{(n)}$ and for each node $i$, the algorithm calculates the value of the class $\Hat{z}_i^{(n+1)}$ that increases $\Hat{l}$ the most, leaving the labels of the other nodes invariant. After updating all the labels of all the nodes in this way, we repeat the update until the new label is identical to the previous one, up to a permutation on the labels. Defining $z[i,p]_j=z_j$ if $j\ne i$ and $z[i,p]_j=p$ otherwise, the algorithm can be summarized as follows: \\

\begin{algorithm}
\caption{The Gradient Descent algorithm}
\label{alg:GD}
\begin{algorithmic}[1]
\Require $z^{(0)}\in\intint{1}{K}^N$
\While {$z^{(n)}\ne z^{(n-1)}$ up to a permutation}
 \For {$i\in \intint{1}{n}$}
    \State $\ z^{(n+1)}_i\gets \argmax_{p\in\intint{1}{N}}$\, $\Hat{l}\left(z^{(n)}[i,p]\right)$
 \EndFor
\EndWhile
\end{algorithmic}
\end{algorithm}
As for Lloyd-type algorithms, we can initialize this algorithm with a noisy version of the real labeling, referred to as the "GD" method, or with the result of the spectral clustering estimation, called the "Spectral+GD" method. \\

\textbf{Variational EM type methods :} This is probably the most widely used method in the literature. Its principle is described in \cite{mariadassou2010uncovering}. A fundamental difference with the previous algorithms is that this algorithm does not aim to estimate the labels of each node. The underlying model actually assumes that the labels are unobserved data, independently drawn according to a distribution $\alpha=(\alpha_p)_{p\in\intint{1}{K}}$ on $\intint{1}{K}$. The idea of the algorithm is therefore first and foremost to estimate $\alpha$ and the probability matrix $P$. In particular, the VEM algorithm computes an estimate of the label distribution $L\left(z\middle|X,\alpha,P\right)$ conditional on the adjacency matrix and other parameters, in the form of a distribution $\rho$ given by
$$
\rho(z)=\prod_{i=1}^N\rho_{iz_i}.
$$
Computing $\Hat{z}_i=\argmax_{p\in\intint{1}{K}}\, \rho_{ip}$, one can have an estimate of the label of each node. This method will called "VEM" if the initialization is random and as "Spectral+VEM" if the initialization is given by the spectral clustering method.

\section{Comparison between "LloydHuber" methods}\label{app:compLloydHuber}

In this section, we compare the performances of "LloydHuber" methods for different values of parameter $r$. Recall that the Huber loss between two vectors $x,y\in\R^d$ is given by \eqref{eq:Huber}. More precisely, we compare the "LloydL1", "LloydL2" and "LloydHuber0.05" methods, where "LloydHuber0.05" corresponds to the LloydHuber" method for $r=0.05$, with a random or a spectral initialization. \\

According to our simulations, the performance in terms of computation time is equivalent for all of these methods. Therefore, we will only present the performance comparison with respect to $b$, $h$ and $\omega$. Also, since the behavior of these methods is similar between the symmetric and asymmetric cases, we will only present the asymmetric case $P^\star=P_{asym}$. The parameter $a$ in the definition of $P_{asym}$ is set to $a=0.9$. \\

\begin{figure}[!ht]
    \centering
    \includegraphics[scale=0.35]{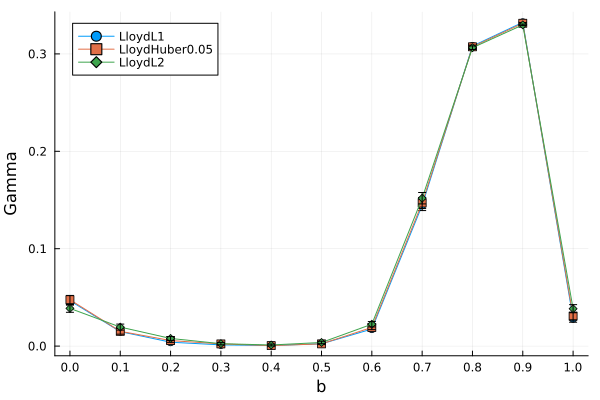}
    \includegraphics[scale=0.35]{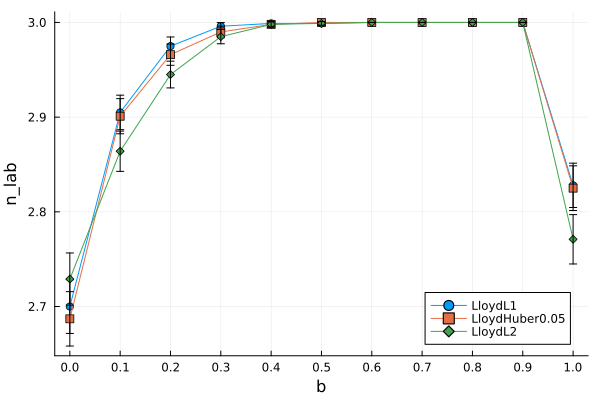}
    \caption{Evolution of $\Gamma$ on the left and of the number of labels recovered on the right according to the identifiability parameter $b\in\{0.0,0.1,\dots,1.0\}$ with $N=50$, $h=0.0$, $a=0.9$ and $\omega=1.0$, for the "LloydHuber" methods randomly initialized.}
    \label{fig:1A}
\end{figure}

\begin{figure}[!ht]
    \centering
    \includegraphics[scale=0.35]{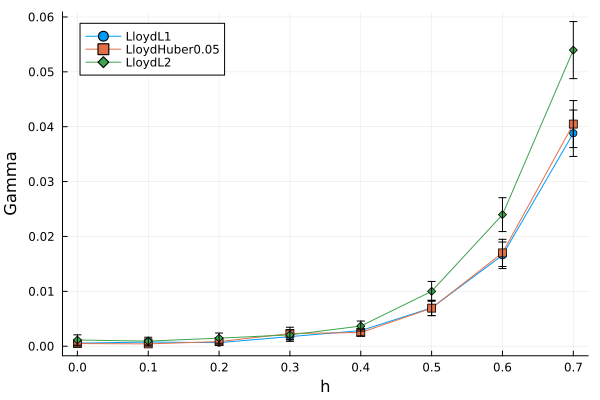}
    \includegraphics[scale=0.35]{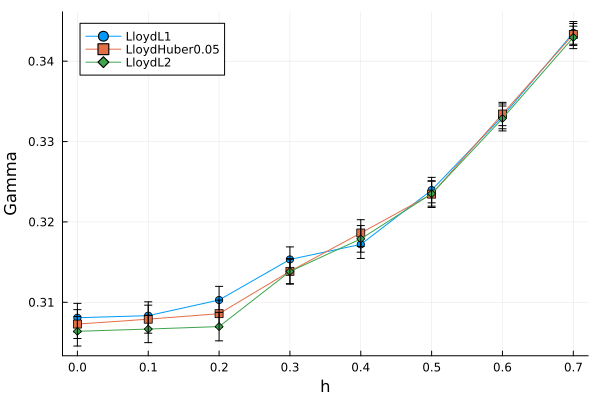}
    \caption{Evolution of $\Gamma$ according to the heterogeneity parameter $h\in\{0.0,0.1,\dots,0.7\}$ with $N=50$ and $\omega=1.0$, for the "LloydHuber" methods randomly initialized in the identifiable case ($a=0.9$, $b=0.4$) on the left and the weakly identifiable case ($a=0.9$, $b=0.8$) on the right.}
    \label{fig:1B}
\end{figure}

\begin{figure}[!ht]
    \centering
    \includegraphics[scale=0.35]{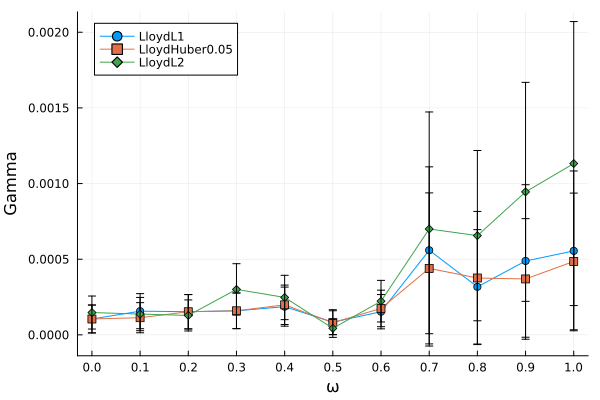}
    \includegraphics[scale=0.35]{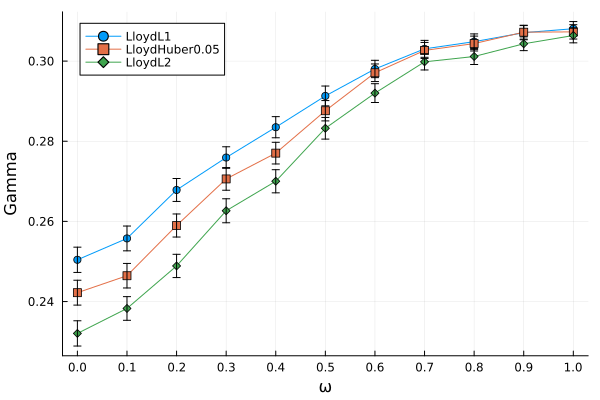}
    \caption{Evolution of $\Gamma$ according to the initial distance parameter $\omega\in\{0.0,0.1,\dots,1.0\}$ with $N=50$ and $h=0.0$ for the "LloydHuber" methods randomly initialized in the identifiable case ($a=0.9$, $b=0.4$) on the left and the weakly identifiable case ($a=0.9$, $b=0.8$) on the right.}
    \label{fig:1C}
\end{figure}

\begin{figure}[!ht]
    \centering
    \includegraphics[scale=0.35]{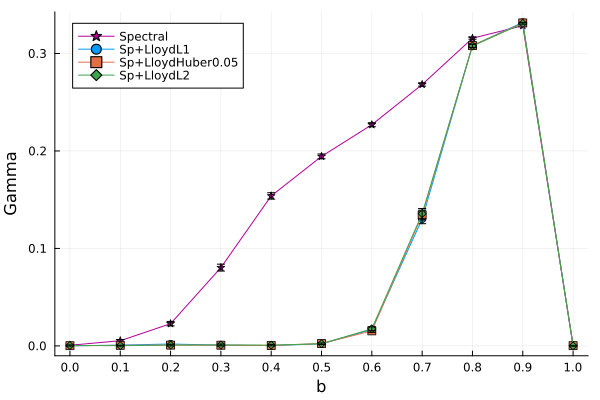}
    \includegraphics[scale=0.35]{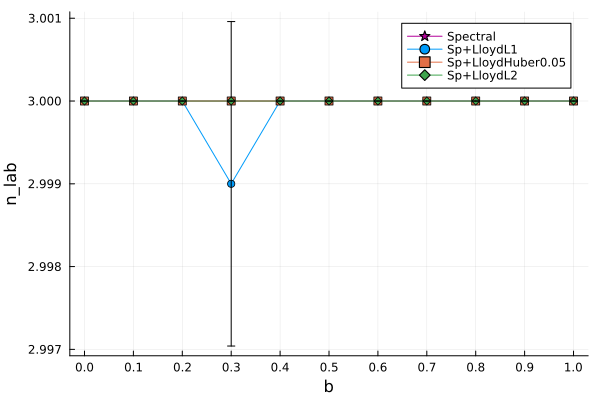}
    \caption{Evolution of $\Gamma$ on the left and of the number of labels recovered on the right according to the identifiability parameter $b\in\{0.0,0.1,\dots,1.0\}$ with $N=50$, $h=0.0$ and $a=0.9$, for the "LloydHuber" methods with spectral initialization.}
    \label{fig:2A}
\end{figure}

\begin{figure}[!ht]
    \centering
    \includegraphics[scale=0.35]{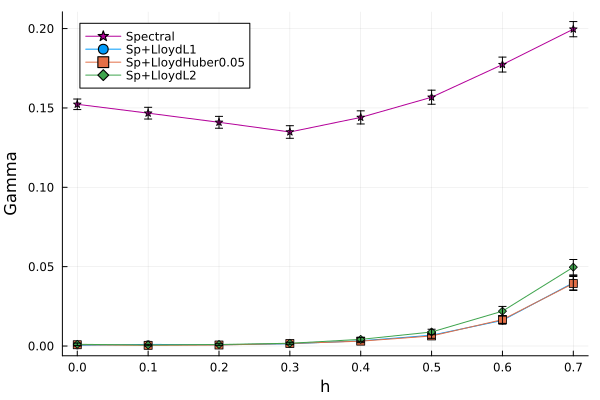}
    \includegraphics[scale=0.35]{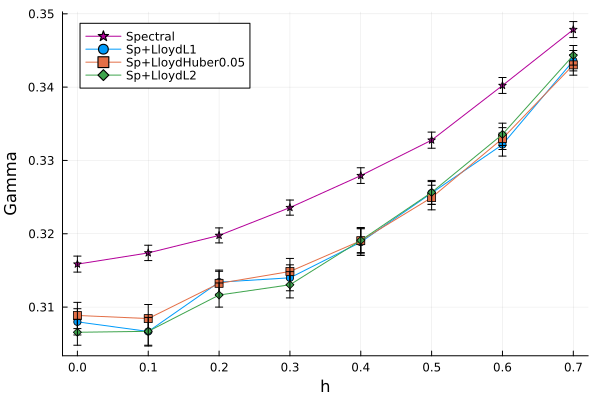}
    \caption{Evolution of $\Gamma$ according to the heterogeneity parameter $h\in\{0.0,0.1,\dots,0.7\}$ with $N=50$, for the "LloydHuber" methods with spectral initialization in the identifiable case ($a=0.9$, $b=0.4$) on the left and the weakly identifiable case ($a=0.9$, $b=0.8$) on the right.}
    \label{fig:2B}
\end{figure}

Figures~\ref{fig:1A}, \ref{fig:1B} and \ref{fig:1C} shows the evolution of $\Gamma$ with respect to $b$, $h$ and $\omega$ respectively for a random initialization. Figures~\ref{fig:2A}, \ref{fig:2B} shows the evolution of $\Gamma$ with respect to $b$ and $h$ respectively for a spectral initialization. Looking at these five figures and Table~\ref{tab:1} and \ref{tab:2}, we can see that the performance of the different algorithms is not significantly different. When initialization is performed with spectral clustering, we find that the performance of all methods is significantly better than that of spectral clustering alone. \\

It should be noted, however, that for a random initialization, the performance of these methods reaches an optimum for $b$ values close to $0.5$ (left-hand side of Figure~\ref{fig:1A}), which seems to contradict the idea that the more identifiable the probability matrix is, the more efficient the inference methods are at finding the true label vector. The right-hand side of Figure~\ref{fig:1A} shows that this maximum is actually explained by the fact that algorithms tend to converge on label vectors containing less than three different labels when $\abs{a-b}$ is large. These minima are avoided if the initial labeling is obtained from the spectral clustering method (right-hand side of Figure~\ref{fig:2B}). Furthermore, Figures~\ref{fig:1B} and \ref{fig:2B} shows that, as with the classical Lloyd algorithm for $k$-means, the more different the class sizes are, the worse the estimate made by the algorithm is. \\

\newpage
\thispagestyle{empty}
~
\newpage

\section{Comparison of different iterative clustering algorithms}\label{app:othermethods}

In this section, the methods we will compare are the "LloydL1", "LloydMLE", "GD" and "VEM". For the sake of brevity, we will only compare these algorithms in the case where initialization is performed by spectral clustering. This choice is motivated by the fact that these algorithms will mainly be used without oracle initialization and that, according to our simulations, the study of this case does not lead to different conclusions to that of random initialization. \\

We will compare the performance with respect to $N$, $b$ and $h$ according to $\Gamma$ and computation time. As previously in Appendix~\ref{app:compLloydHuber}, we will only present the asymmetric case $P^\star=P_{asym}$ since the comparison in the symmetric case leads to the same conclusions. The parameter $a$ in the definition of $P_{asym}$ is set to $a=0.9$. \\

\begin{figure}[!ht]
    \centering
    \includegraphics[scale=0.35]{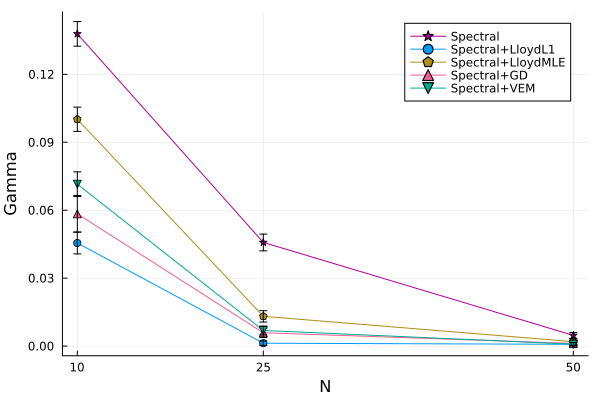}
    \includegraphics[scale=0.35]{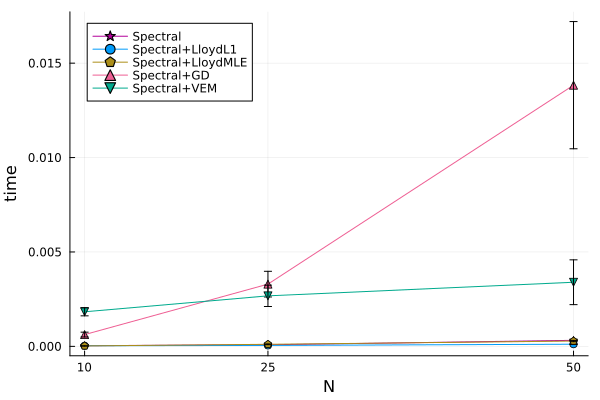}
    \caption{Evolution of $\Gamma$ on the left and the mean computation time on the right according to the number of nodes $N\in\{10,25,50\}$ with $a=0.9$, $b=0.1$ and $h=0.0$, for the "LloydL1", "LloydMLE", "GD" and "VEM" methods with spectral initialization.}
    \label{fig:4C}
\end{figure}

\begin{figure}[!ht]
    \centering
    \includegraphics[scale=0.35]{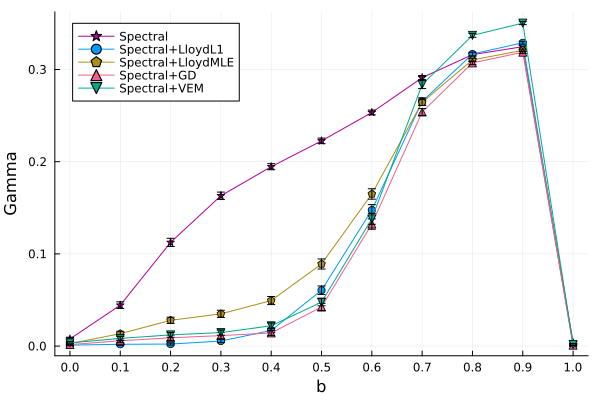}
    \includegraphics[scale=0.35]{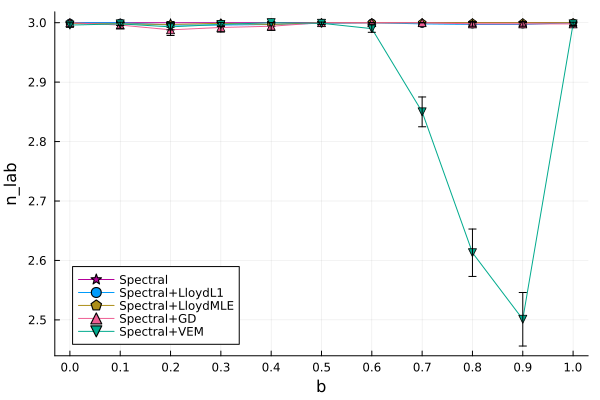}
    \caption{Evolution of $\Gamma$ on the left and of the number of labels recovered on the right according to the identifiability parameter $b\in\{0.0,0.1,\dots,1.0\}$ with $N=25$, $a=0.9$ and $h=0.0$, for the "LloydL1", "LloydMLE", "GD" and "VEM" methods with spectral initialization.}
    \label{fig:4A}
\end{figure}

Figures~\ref{fig:4C} clearly shows that the “LloydL1” is both the best in the case where there are few nodes and the one with the lowest computation time. Note that this figure also highlights the fact that the classically used VEM method becomes very slow when the number of nodes is large. Figures~\ref{fig:4A} shows that the performance of the $4$ methods is relatively similar when $N$ is large, with a clear improvement over using the Spectral method alone. Finally, the last figure shows that the “VEM” method performs slightly better in relation to class size heterogeneity.

\begin{figure}[!ht]
    \centering
    \includegraphics[scale=0.35]{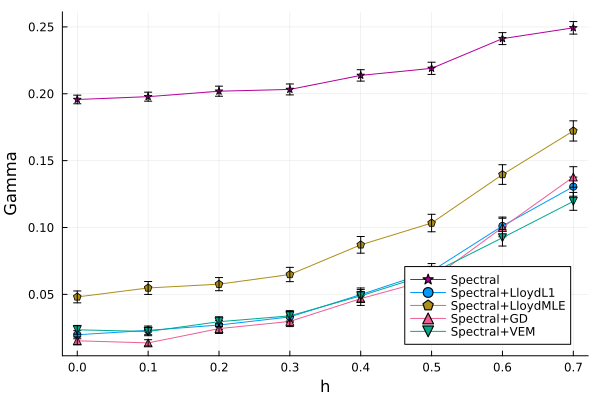}
    \includegraphics[scale=0.35]{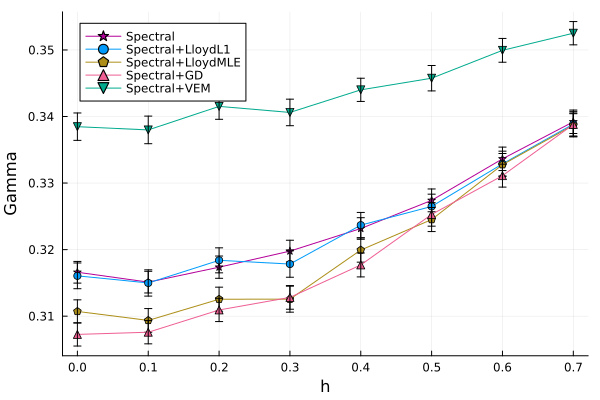}
    \caption{Evolution of $\Gamma$ according to the heterogeneity parameter $h\in\{0.0,0.1,\dots,0.7\}$ with $N=25$, for the "LloydL1", "LloydMLE", "GD" and "VEM" methods with spectral initialization in the identifiable case ($a=0.9$, $b=0.4$) on the left and the weakly identifiable case ($a=0.9$, $b=0.8$) on the right.}
    \label{fig:4B}
\end{figure}

\newpage

\begin{table}[!ht]
    \centering
    \begin{tabular}{cccc|ccc|ccc}
	   $P$ & $h$ & $b$ & $\omega$ & $\Gamma$ & & & time & & \\
	   & & & & LloydL1 & LloydHuber0.05 & LloydL2 & LloydL1 & LloydHuber0.05 & LloydL2 \\
	   \hline
	   $P_{asym}$ & 0.0 & 0.4 & 0.2 & 0.0 & 0.0 & 0.0 & 0.0 & 0.0 & 0.0 \\
	 & & & 1.0 & 0.001 & 0.0 & 0.001 & 0.0 & 0.0 & 0.0 \\
	 & & 0.8 & 0.2 & 0.268 & 0.259 & 0.249 & 0.003 & 0.003 & 0.003 \\
	 & & & 1.0 & 0.308 & 0.307 & 0.306 & 0.003 & 0.003 & 0.003 \\
	 & 0.7 & 0.4 & 0.2 & 0.033 & 0.033 & 0.041 & 0.0 & 0.0 & 0.0 \\
	 & & & 1.0 & 0.039 & 0.04 & 0.054 & 0.0 & 0.0 & 0.0 \\
	 & & 0.8 & 0.2 & 0.328 & 0.325 & 0.324 & 0.003 & 0.004 & 0.004 \\
	 & & & 1.0 & 0.344 & 0.343 & 0.343 & 0.003 & 0.003 & 0.003 \\
      \hline
	   $P_{sym}$ & 0.0 & 0.4 & 0.2 & 0.0 & 0.0 & 0.0 & 0.0 & 0.0 & 0.0 \\
	 & & & 1.0 & 0.023 & 0.029 & 0.03 & 0.0 & 0.001 & 0.0 \\
	 & & 0.8 & 0.2 & 0.272 & 0.263 & 0.249 & 0.003 & 0.004 & 0.004 \\
	 & & & 1.0 & 0.315 & 0.316 & 0.316 & 0.003 & 0.003 & 0.003 \\
	 & 0.7 & 0.4 & 0.2 & 0.006 & 0.005 & 0.008 & 0.0 & 0.0 & 0.0 \\
	 & & & 1.0 & 0.082 & 0.084 & 0.087 & 0.001 & 0.001 & 0.001 \\
	 & & 0.8 & 0.2 & 0.28 & 0.277 & 0.279 & 0.003 & 0.003 & 0.003 \\
	 & & & 1.0 & 0.32 & 0.321 & 0.324 & 0.003 & 0.003 & 0.003 \\
    \end{tabular}
    \caption{Comparison between "LloydL1", "LloydHuber0.05" and "LloydL2" methods with random initialization.}
    \label{tab:1}
\end{table}

\begin{table}[!ht]
    \centering
    \begin{tabular}{ccc|cccc} 
        $P$ & $h$ & $b$ & $\Gamma$ & & & \\
        & & & Spectral & Sp+LloydL1 & Sp+LloydHuber0.05 & Sp+LloydL2 \\
	    \hline
	    $P_{asym}$ & 0.0 & 0.4 & 0.154 & 0.001 & 0.001 & 0.0 \\
	    & & 0.8 & 0.316 & 0.308 & 0.308 & 0.307 \\
	    & 0.7 & 0.4 & 0.199 & 0.039 & 0.039 & 0.05 \\
	    & & 0.8 & 0.348 & 0.343 & 0.344 & 0.344 \\
	    \hline
	    $P_{sym}$ &	0.0 & 0.4 & 0.006 & 0.001 & 0.001 & 0.002 \\
	    & & 0.8 & 0.314 & 0.309 & 0.309 & 0.309 \\
	    & 0.7 & 0.4 & 0.086 & 0.026 & 0.025 & 0.029 \\
	    & & 0.8 & 0.334 & 0.312 & 0.314 & 0.314 \\
\end{tabular}
    \caption{Comparison between "Spectral" and "Spectral+LloydL1", "Spectral+LloydHuber0.05" and "Spectral+LloydL2" algorithms (refer to as Spectral, Sp+LloydL1, Sp+LloydHuber0.05 and Sp+LloydL2 in the table).}
    \label{tab:2}
\end{table}

\begin{table}[!ht]
    \centering
    \begin{tabular}{cccc|ccccc}
	 $P$ & $N$ & $h$ & $b$ & $\Gamma$ & & & & \\
	 & & & & Spectral & Sp+LloydL1 & Sp+LloydMLE & Sp+GD & Sp+VEM \\
	\hline
	$P_{asym}$ & 10 & 0.0 & 0.4 & 0.221 & 0.166 & 0.191 & 0.167 & 0.177 \\
	 & & & 0.8 & 0.303 & 0.299 & 0.292 & 0.336 & 0.317 \\
	 & & 0.7 & 0.4 & 0.247 & 0.212 & 0.228 & 0.23 & 0.219 \\
	 & & & 0.8 & 0.309 & 0.311 & 0.304 & 0.334 & 0.324 \\
	 & 25 & 0.0 & 0.4 & 0.196 & 0.02 & 0.051 & 0.013 & 0.022 \\ 
	 & & & 0.8 & 0.317 & 0.317 & 0.31 & 0.308 & 0.339 \\
	 & & 0.7 & 0.4 & 0.252 & 0.133 & 0.173 & 0.141 & 0.122 \\
	 & & & 0.8 & 0.339 & 0.338 & 0.339 & 0.339 & 0.352 \\
	 & 50 & 0.0 & 0.4 & 0.151 & 0.001 & 0.009 & 0.004 & 0.006 \\
	 & & & 0.8 & 0.317 & 0.307 & 0.304 & 0.3 & 0.341 \\
	 & & 0.7 & 0.4 & 0.201 & 0.04 & 0.038 & 0.03 & 0.027 \\
	 & & & 0.8 & 0.349 & 0.344 & 0.346 & 0.346 & 0.357 \\
	 \hline
	 $P_{sym}$ & 10 & 0.0 & 0.4 & 0.183 & 0.135 & 0.143 & 0.169 & 0.125 \\
	 & & & 0.8 & 0.288 & 0.285 & 0.283 & 0.322 & 0.3 \\
	 & & 0.7 & 0.4 & 0.196 & 0.153 & 0.154 & 0.182 & 0.126 \\
	 & & & 0.8 & 0.302 & 0.296 & 0.295 & 0.322 & 0.307 \\
	 & 25 & 0.0 & 0.4 & 0.072 & 0.016 & 0.013 & 0.01 & 0.016 \\
	 & & & 0.8 & 0.309 & 0.308 & 0.304 & 0.306 & 0.333 \\
	 & & 0.7 & 0.4 & 0.14 & 0.065 & 0.058 & 0.061 & 0.036 \\
	 & & & 0.8 & 0.331 & 0.324 & 0.323 & 0.324 & 0.338 \\
	 & 50 & 0.0 & 0.4 & 0.005 & 0.002 & 0.001 & 0.001 & 0.001 \\
	 & & & 0.8 & 0.314 & 0.309 & 0.305 & 0.306 & 0.338 \\
	 & & 0.7 & 0.4 & 0.09 & 0.024 & 0.02 & 0.028 & 0.018 \\
	 & & & 0.8 & 0.334 & 0.313 & 0.307 & 0.305 & 0.318 \\
\end{tabular}
    \caption{Comparison between "Spectral" and "Spectral+LloydL1", "Spectral+LloydMLE", "Spectral+GD" and "Spectral+VEM" algorithms (refer to as Spectral, Sp+LloydL1, Sp+LloydMLE, Sp+GD and SP+VEM in the table) according to $\Gamma$.}
    \label{tab:3}
\end{table}

\begin{table}[!ht]
    \centering
    \begin{tabular}{cccc|ccccc}
	 $P$ & $N$ & $h$ & $b$ & time & & & & \\
	 & & & & Spectral & Sp+LloydL1 & Sp+LloydMLE & Sp+GD & Sp+VEM \\
	\hline
	$P_{asym}$ & 10 & 0.0 & 0.4 & 0.0 & 0.0 & 0.0 & 0.003 & 0.004 \\
	 & & & 0.8 & 0.0 & 0.0 & 0.0 & 0.003 & 0.006 \\
	 & & 0.7 & 0.4 & 0.0 & 0.0 & 0.0 & 0.003 & 0.005 \\
	 & & & 0.8 & 0.0 & 0.0 & 0.0 & 0.003 & 0.006 \\
	 & 25 & 0.0 & 0.4 & 0.0 & 0.0 & 0.0 & 0.008 & 0.009 \\
	 & & & 0.8 & 0.0 & 0.001 & 0.001 & 0.084 & 0.064 \\
	 & & 0.7 & 0.4 & 0.0 & 0.0 & 0.0 & 0.034 & 0.026 \\
	 & & & 0.8 & 0.0 & 0.001 & 0.001 & 0.084 & 0.062 \\
	 & 50 & 0.0 & 0.4 & 0.0 & 0.0 & 0.001 & 0.029 & 0.014 \\
	 & & & 0.8 & 0.0 & 0.002 & 0.007 & 0.658 & 0.356 \\
	 & & 0.7 & 0.4 & 0.0 & 0.0 & 0.001 & 0.069 & 0.042 \\
	 & & & 0.8 & 0.0 & 0.003 & 0.008 & 0.678 & 0.338 \\
	 \hline
	 $P_{sym}$ & 10 & 0.0 & 0.4 & 0.0 & 0.0 & 0.0 & 0.003 & 0.003 \\
	 & & & 0.8 & 0.0 & 0.0 & 0.0 & 0.003 & 0.005 \\
	 & & 0.7 & 0.4 & 0.0 & 0.0 & 0.0 & 0.003 & 0.003 \\
	 & & & 0.8 & 0.0 & 0.0 & 0.0 & 0.003 & 0.005 \\
	 & 25 & 0.0 & 0.4 & 0.0 & 0.0 & 0.0 & 0.006 & 0.006 \\
	 & & & 0.8 & 0.0 & 0.001 & 0.001 & 0.075 & 0.062 \\
	 & & 0.7 & 0.4 & 0.0 & 0.0 & 0.0 & 0.02 & 0.015 \\
	 & & & 0.8 & 0.0 & 0.001 & 0.001 & 0.077 & 0.064 \\
	 & 50 & 0.0 & 0.4 & 0.0 & 0.0 & 0.0 & 0.012 & 0.004 \\
	 & & & 0.8 & 0.0 & 0.002 & 0.007 & 0.579 & 0.311 \\
	 & & 0.7 & 0.4 & 0.0 & 0.0 & 0.001 & 0.06 & 0.042 \\
	 & & & 0.8 & 0.0 & 0.002 & 0.006 & 0.574 & 0.355 \\
\end{tabular}
    \caption{Comparison between "Spectral" and "Spectral+LloydL1", "Spectral+LloydMLE", "Spectral+GD" and "Spectral+VEM" algorithms (refer to as Spectral, Sp+LloydL1, Sp+LloydMLE, Sp+GD and SP+VEM in the table) according to the computation time.}
    \label{tab:4}
\end{table}

\end{document}